# Response Aware Model-Based Collaborative Filtering

Guang Ling[1,2], Haiqin Yang[1,2], Michael R. Lyu[1,2], and Irwin King[2,3]

[1]Shenzhen Research Institute, The Chinese University of Hong Kong, Shenzhen, China
[2]Department of Computer Science and Engineering, The Chinese University of Hong Kong,
Shatin, N.T., Hong Kong {gling,hqyang,lyu,king}@cse.cuhk.edu.hk
[3]AT&T Labs Research, San Francisco, CA, USA, irwin@research.att.com

## Abstract

Previous work on recommender systems mainly focus on fitting the ratings provided by users. However, the response patterns, i.e., some items are rated while others not, are generally ignored. We argue that failing to observe such response patterns can lead to *biased* parameter estimation and sub-optimal model performance. Although several pieces of work have tried to model users' response patterns, they miss the effectiveness and interpretability of the successful matrix factorization collaborative filtering approaches. To bridge the gap, in this paper, we unify explicit response models and PMF to establish the Response Aware Probabilistic Matrix Factorization (RAPMF) framework. We show that RAPMF subsumes PMF as a special case. Empirically we demonstrate the merits of RAPMF from various aspects.

## 1 Introduction

Recently, online music and video streaming services have seen an explosive growth. As the user base and contents expanding tremendously, recommender systems become crucial for service providers. Cloud-based music streaming services such as iTunes Match, Google Music, Yahoo! Music, Pandora, Songify, etc. make it easier than ever to rate songs and buy new music. With the rocketing growth of the number of users, their explicit ratings become more and more accessible. Effective usage of these ratings can lead to high quality recommendation, which is vital for the cloud-based online streaming services. This is because most services charge very little subscription fee, if not none. The main income comes from the selling of music. Nowadays, online streaming services often have a large user base, so even if a small change in recommendation quality may have dramatic effect on sales.

Due to immense market value, various recommendation techniques have been proposed. Generally, these approaches can be classified into neighborhood-based methods and model-based methods. Typical neighborhood-based approaches includes user-based methods [1, 4] and item-based methods [3, 11, 25]. The state-of-the-art model-based methods include restricted Boltzmann machines [24], SVD++ [8, 9], Probabilistic Matrix Factorization (PMF) [22], and multi-domain collaborative filtering [32], graphical models [7], pair-wise tensor factorization [21], and matrix factorization with social regularization [14], etc.

However, in real-world rating systems, users' ratings carry twofold information. Firstly the rating value indicates a user's preference on a particular item as well as an item's inherent features. The scores that a user assigns to different items convey information on what the user likes and what the user dislikes. The rating values that an item received from different users also carry information on intrinsic properties of the item. Second, the ratings also reveal users' response patterns, i.e., some items are rated while others not. This information can be utilized to improve the model performance. However, previously proposed methods usually assume that all the users would rate *all* the inspected items, or more generally, *randomly* select inspected items to rate. These methods fit the users' ratings directly and ignore the key factor, users' response patterns. The ignorance will degrade the model performance. In this paper, we explore previously ignored response information to further boost recommender system's quality.

Practically, the assumption of *all inspection* or *randomly rate* is not true in real-world rating systems. Users are unlikely to rate all the inspected items or randomly select the inspected items to rate. Shown in Figure 1(a) is the rating value distribution of the items that users *choose to rate*, while Figure 1(b) shows the distribution of ratings for *randomly* selected songs from the same group of users. Clearly these two distributions are very different. In the user selected ratings, there are far more items with high ratings than that in the randomly selected songs. This is compelling evidence showing that the assumption that all the users would rate all the inspected items or select random items to rate is unlikely to be true. The investigation of the Yahoo!LaunchCast data indicates that users are more likely to rate items they do love and hate, but not neutral [16, 26].

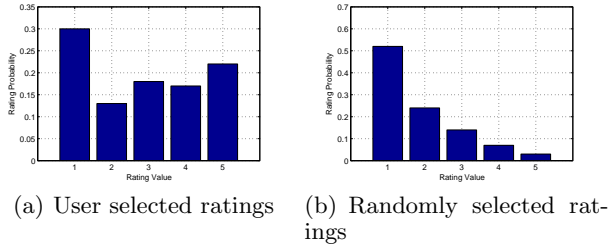

(a) User selected ratings   (b) Randomly selected ratings

Figure 1: Distribution of ratings in a music website [16].

To further demonstrate the risk of *incorrect* parameter estimation and *biased* rating prediction when ignoring the response information, we show an intuitive example of five users' rating on five items in Table 1, where the ratings are skewed to either 4 or 5. Clearly, user-based approaches [1, 4] and item-based approaches [3, 11, 25] are more likely to predict rating values in the range of 4 to 5. Similarly, ignoring response information will cause the one-class issue for model-based approaches [10, 17, 18]. In a real-world recommender system, the case may not be as extreme as is in Table 1. Nevertheless, the effect is similar. By ignoring the response information, we will learn a model that has *bias*.

Currently, there are two main streams of work trying to solve the above response ignorance problem. One line of work try to model the above phenomena as a one-class collaborative filtering task [10, 17, 18]. A heuristic weight in the range of 0 to 1 is introduced to calibrate the loss on those unseen ratings, where the rating scores are set to zeros [17, 18]. Embedding user information is also adopted to optimize the weight on the unseen ratings via users' similarity [10]. However, these methods do not model the users' missing response information together with the ratings. The other line of work model the response ignorance through missing data theory [13]. The multinomial mixture model is adopted to model the non-random response [16]. The work is also extended for collaborative ranking [15]. These methods model users' response patterns and ratings via multinomial mixture model, but they discard the effectiveness and interpretability of the matrix factorization approaches [8, 22].

To bridge this gap, we are the first to integrate the users' response patterns into PMF to establish a unified framework, which we refer to as Response Aware PMF (RAPMF). The response models we propose include the rating dominating response model, and a generalized one, the context-aware response model. We demonstrate the advantages of our proposed RAPMF through detailed and fair experimental comparison.

The rest of the paper is organized as follows. In Section 2, we motivate the explicit modeling of user responses from a probabilistic point of view. In Section 3, we present how to incorporate response models into PMF and elaborate the proposed RAPMF model. Empirical study and comparison with previous work is conducted in Section 4. The paper is concluded in Section 5.

## 2 Response and Missing Theory

Modeling response patterns have a strong incentive from statistical missing data theory [13]. The response patterns can be hidden [2, 6] or explicit. In recommender system case, it is explicit. In the following, we show that without modeling the response patterns properly, we may learn a *bias* model.

### 2.1 Setup and Notation

Assume that we are given a partially observed $N \times M$ matrix $X$, where $N$ is the number of users and $M$ is the number of items, the $(i, j)$ element of $X$ denotes the rating assigned by user $i$ to item $j$ in the scale of 1 to $D$. Collaborative filtering approaches try to recover the original full matrix $X_{full}$ to predict users' preferences.

In the matrix $X$, an unobserved entry is denoted by 0. Alternatively, we denote all the observations as a set of triplets $(i, j, x) \in \mathcal{Q}$. Moreover, we define a companion response indicator matrix $R$ to denote whether the corresponding rating is observed in $X$. If $X_{ij} \neq 0$, i.e., we have observed user $i$'s rating on item $j$, then $R_{ij} = 1$. Otherwise $R_{ij} = 0$. Note that $X$ is partially observed while $R$ is fully observed.

Table 1: Skewed ratings on 5 items from 5 users

|       | item1 | item2 | item3 | item4 | item5 |
|-------|-------|-------|-------|-------|-------|
| user1 | 5     | 4     |       |       |       |
| user2 |       | 5     |       | 4     |       |
| user3 | 4     |       |       | 4     |       |
| user4 | 5     |       | 5     |       |       |
| user5 |       | 4     |       |       | 5     |

## 2.2 Missing Data Theory

Following missing data theory in [13], we model the collaborative filtering data as a two-step procedure. First, a data model $P(X|\theta)$ generates the full data matrix $X_{full}$. Then, a response model $P(R|X,\mu)$ determines which elements in $X_{full}$ are observed. Hence, we can take a parametric joint distribution on the observed data matrix $X$ and the response matrix $R$, conditioned on the model parameters, $\theta$ and $\mu$.

$$P(R, X|\mu, \theta) = P(R|X, \mu, \theta)P(X|\mu, \theta) \quad (1)$$
$$= P(R|X, \mu)P(X|\theta), \quad (2)$$

where $P(R|X,\mu)$ is also referred to as the missing data model. In the following, we use response model and missing data model interchangeably.

According to the missing data theory [13], there are three kinds of missing data assumptions: 1) Missing Completely At Random (MCAR); 2) Missing At Random (MAR), and 3) Not Missing At Random (NMAR). MCAR has the strongest independence assumption. Under the MCAR assumption, the missing mechanism cannot depend on the data in any way. Whether we will observe a response is fully determined by the parameter $\mu$ and is irrelevant to the users' rating, i.e.,

$$P(R|X, \mu) = P(R|\mu) \quad (3)$$

One typical example where MCAR holds is that given an inspected item, whether it will be observed is a Bernoulli trail with probability $\mu$.

The MAR assumption is slightly different from the MCAR assumption. Let $X_{full} = (X_{obs}, X_{mis})$, i.e., the full data matrix $X_{full}$ is separated into observed data matrix $X_{obs}$ and missing data matrix $X_{mis}$. Under the MAR assumption, the response probability depends on the *observed* data and $\mu$, i.e.,

$$P(R|X, \mu) = P(R|X_{obs}, \mu). \quad (4)$$

Marlin and Zemel [15] refer to this as the probability of observing a particular response only depending on the observed elements of the data vector. The assumption made by MAR may seem bizarre. However, it comes up naturally if we want to ignore response model and still learn *unbiased* data model parameters. We demonstrate this in the following.

Let $\mathcal{L}(\mu, \theta|X_{obs}, R)$ be the likelihood of $\mu$ and $\theta$ given the observation $X_{obs}$ and $R$. Under the MAR assumption, we have

$$\begin{aligned}
\mathcal{L}(\mu, \theta|X_{obs}, R) &= P(R, X_{obs}|\mu, \theta) \\
&= \int_{X_{mis}} P(R, X|\mu, \theta) dX_{mis} \\
&= \int_{X_{mis}} P(R|X, \mu) P(X|\theta) dX_{mis} \\
&= \int_{X_{mis}} P(R|X_{obs}, \mu) P(X|\theta) dX_{mis} \quad (5) \\
&= P(R|X_{obs}, \mu) \int_{X_{mis}} P(X|\theta) dX_{mis} \\
&= P(R|X_{obs}, \mu) P(X_{obs}|\theta) \\
&\propto P(X_{obs}|\theta).
\end{aligned}$$

The key to marginalize the missing data is that the missing data model depends only on the observed data, i.e., MAR assumption in Eq. (4). Under the MCAR assumption, we can simplify Eq. (5) similarly, which only depends on $\mu$. Note that the assumption made by MAR appears naturally in the derivation. This is the independence assumption we cannot release anymore without losing the ability to marginalize the complete data model independently of missing data model.

If both MCAR and MAR fail to hold, then NMAR assumption is made. Unlike MCAR and MAR, NMAR requires an explicit response model in order to learn unbiased model parameters. Otherwise, maximizing $P(X_{obs}|\theta)$ directly can yield a *biased* $\theta$. With only a few exceptions [15, 16], nearly all the previous work on recommender systems try to maximize the data model directly [5, 14, 19, 22, 23]. In a typical recommender system, the data collected can easily violate the MAR assumption. The distinct distribution of rating values on user selected items and randomly selected items hints that the response pattern depends on not only the observed data. Also, a survey on the Yahoo! LanuchCast provides evidence that the response probability might depend on the fondness of particular items [16, 26].

## 3 Models and Analysis

In the following, we first review the Probabilistic Matrix Factorization (PMF). After that, we present the response aware PMF and show how it can incorporate PMF with the response models. More specifically, we introduce two response models, the rating dominant response model and the context-aware response model. The updating rules and complexity analysis are provided correspondingly.

## 3.1 Probabilistic Matrix Factorization

PMF [22] is one of the most famous matrix factorization models in collaborative filtering, which decomposes the partially observed data matrix $X$ into the product of two low-rank latent feature matrices, $U$ and $V$, where $U \in \mathbb{R}^{K \times N}$, $V \in \mathbb{R}^{K \times M}$, and $K \ll \min(N, M)$.

By assuming Gaussian distribution on the residual noise of observed data and placing Gaussian priors on the latent feature matrices, PMF tries to maximize the log-likelihood of the posterior distribution on the user and item features as follows:

$$\mathcal{L}_{PMF} = - \sum_{(i,j,x) \in \mathcal{Q}} \frac{(x - U_i^T V_j)^2}{2\sigma^2} - \frac{\|U\|_F^2}{2\sigma_U^2} - \frac{\|V\|_F^2}{2\sigma_V^2}. \tag{6}$$

This is equivalent to minimizing a squared loss with regularization defined as follows:

$$\mathcal{E} = \frac{1}{2} \sum_{(i,j,x) \in \mathcal{Q}} (x - U_i^T V_j)^2 + \frac{\lambda_U}{2}\|U\|_F^2 + \frac{\lambda_V}{2}\|V\|_F^2, \tag{7}$$

where $\lambda_U = \sigma^2/\sigma_U^2$ and $\lambda_V = \sigma^2/\sigma_V^2$ are positive constants to control the trade-off between the loss and the regularization terms. $\|\cdot\|_F^2$ denotes the Frobenious norm.

After training the PMF model via gradient descent or stochastic gradient algorithms [22], the predicted rating that user $i$ would assign to item $j$ can be computed as the expected mean of the Gaussian distribution $\hat{x}_{ij} = U_i^T V_j$.

## 3.2 Response Aware PMF

In Sec. 2, we have demonstrated that by neglecting response patterns, not only do we lose the potential information that might boost the model performance, but also can it lead to incorrect or biased parameters estimation. Due to the effectiveness and interpretability of PMF, we will unify it with explicit response models, which we refer to as Response Aware PMF (RAPMF).

Replacing $\theta$ in Eq. (2) by the low-rank latent feature matrices in PMF, we have

$$P(R, X | U, V, \mu, \sigma^2) = P(R | X, U, V, \mu, \sigma^2) P(X | U, V, \sigma^2). \tag{8}$$

The probability of full model, $P(R, X | U, V, \mu, \sigma^2)$, is decomposed into data model $P(X | U, V, \sigma^2)$ and the missing data model $P(R | X, U, V, \mu, \sigma^2)$.

## 3.3 Response Model

Modeling the missing data successfully requires a correct and tractable distribution on the response patterns. Bernoulli distribution is an intuitive distribution to explain data missing phenomena [16]. Depending on whether users' and items' features are incorporated, we propose two response models, *rating dominant response model* and *context-aware response model*.

## 3.4 Rating Dominant Response Model

For the sake of simplification, we assume the probability that a user chooses to rate an item follows a Bernoulli distribution given the rating assigned is $k$. Hence, for a scale of 1 to $D$, the rating dominant response model has $D$ parameters $\mu_1, \mu_2, \cdots, \mu_D$.

If $X$ is fully observed, then the response mechanism can be modeled as [16]:

$$P(R|X, U, V, \mu, \sigma^2) = P(R|X, \mu)$$
$$= \prod_{i=1}^{N} \prod_{j=1}^{M} \prod_{k=1}^{D} (\mu_k^{[r_{ij}=1]} (1-\mu_k)^{[r_{ij}=0]})^{[x_{ij}=k]}, \tag{9}$$

where $[r = 0]$ is an indicator variable that outputs 1 if the expression is valid and 0 otherwise. It is noted that Eq. (9) adopts the "winner-take-all" scheme, i.e., a hard assignment scheme, to model users' response on a particular rating.

However, in real-world recommender system, the data is not fully observed. The "winner-take-all" scheme brings the risk of deteriorating assignment probability when the data is recovered based on the learned model. Hence, we adopt a soft assignment using probability of the possible rating values in the response model as follows:

$$P(R|X, U, V, \mu, \sigma^2) = P(R|U, V, \mu, \sigma^2)$$
$$= \prod_{i=1}^{N} \prod_{j=1}^{M} \sum_{k=1}^{D} (\mu_k^{[r_{ij}=1]} (1-\mu_k)^{[r_{ij}=0]}) P(x_{ij} = k | U, V, \sigma^2), \tag{10}$$

where $P(X|U, V, \sigma^2)$, the probability of $X$ being assigned to $k$, can be set to $\mathcal{N}(k|U^T V, \sigma^2)$ as is in [22].

To relieve the inaccuracy issue when recovering the original model, we further introduce a discount parameter $\beta$ on the assignment probability

$$P(R|X, U, V, \mu, \sigma^2) = P(R|U, V, \mu, \sigma^2) \tag{11}$$
$$\propto \prod_{i=1}^{N} \prod_{j=1}^{M} (\sum_{k=1}^{D} (\mu_k^{[r_{ij}=1]} (1-\mu_k)^{[r_{ij}=0]}) \mathcal{N}(k|U^T V, \sigma^2))^\beta, \tag{12}$$

where the parameter $\beta$, in the range of 0 to 1, can be interpreted as the faith we have on the response model relative to the data model. As $\beta$ decreases, the effect of the response model decreases correspondingly. When $\beta = 0$, the RAPMF collapses to PMF.

More importantly, the expectations of Bernoulli distributions, $\mu_k$'s should be in the range of 0 to 1. With the performance consideration, the logistic function is usually adopted to constrain the range of $\mu_k$'s [15],

$$g(\mu_k) = \frac{1}{1 + \exp(-\mu_k)}, \quad k = 1, \ldots, D. \quad (13)$$

Similarly, we place a zero mean Gaussian prior on $\mu_k$ to regularize it.

Note that in Eq. (10), we use only one parameter for each possible rating value, so all the users and items share the same probability as long as the rating values are the same. This is a simple approach to capture the intuition that the rating assigned to an item may influence the chance that it got rated. This motivates us to name this model as *rating dominant response model*. We refer to PMF with Rating dominate response model as RAPMF-r.

By incorporating the response model in Eq. (12) and the PFM model in Eq. (6) into RAPMF in Eq. (8), we obtain the log-likelihood of the RAPMF-r as follows:

$$\mathcal{L}(U, V, \sigma^2, \mu)$$
$$= \beta \sum_{i=1}^{N} \sum_{j=1}^{M} \log(\sum_{k=1}^{D} \alpha_{kij} \mathcal{N}(k|U^T V, \sigma^2)) - \frac{1}{2\sigma_\mu^2}\|\mu\|^2 -$$
$$\sum_{(i,j,x) \in Q} \frac{(x_{ij} - U_i^T V_j)^2}{2\sigma^2} - \frac{1}{2\sigma_U^2}\|U\|_F^2 - \frac{1}{2\sigma_V^2}\|V\|_F^2 + C, \quad (14)$$

where $C$ denotes the constant terms and $\alpha_{kij}$ is defined as

$$\alpha_{kij} = (g(\mu_k)^{[r_{ij}=1]}(1 - g(\mu_k))^{[r_{ij}=0]}). \quad (15)$$

The gradient of $\mathcal{L}$ with respect to $U_i$ is:

$$\frac{\partial \mathcal{L}}{\partial U_i} = -\beta \sum_{j=1}^{M} \frac{\sum_{k=1}^{D} \alpha_{kij} \mathcal{N}(k|U^T V, \sigma^2)(U_i^T V_j - k) V_j}{\sum_{k=1}^{D} \alpha_{kij} \mathcal{N}(k|U^T V, \sigma^2)}$$
$$- \sum_{j=1}^{M} (U_i^T V_j - x_{ij})[r_{ij} = 1] V_j - \lambda_U U_i. \quad (16)$$

Similarly, the gradient of $\mathcal{L}$ with respect to $V_j$ is:

$$\frac{\partial \mathcal{L}}{\partial V_j} = -\beta \sum_{i=1}^{N} \frac{\sum_{k=1}^{D} \alpha_{kij} \mathcal{N}(k|U^T V, \sigma^2)(U_i^T V_j - k) U_i}{\sum_{k=1}^{D} \alpha_{kij} \mathcal{N}(k|U^T V, \sigma^2)}$$
$$- \sum_{j=1}^{M} (U_i^T V_j - x_{ij})[r_{ij} = 1] U_i - \lambda_V V_j. \quad (17)$$

Both Eq. (16) and Eq. (17) consist of three terms. The first term corresponds to the change due to the response model, the second term is the change due to the data model and third is a regularization to avoid overfitting. Note that by adjusting $\beta$, we effectively alter the weight of the response model when updating parameters.

Finally, the gradient of $\mathcal{L}$ with respect to $\mu_l$ is

$$\frac{\partial \mathcal{L}}{\partial \mu_l} = \sum_{i=1}^{N} \sum_{j=1}^{M} \frac{\mathcal{N}(l|U^T V, \sigma^2) g'(\mu_l)(-1)^{[r_{ij}=0]}}{\sum_{k=1}^{D} \alpha_{kij} \mathcal{N}(k|U^T V, \sigma^2)} - \lambda_\mu \mu_l, \quad (18)$$

where $g'(x)$ is the derivative of the sigmoid function $g(x)$. In Eq. (16), (17), (18), $\lambda_U = \sigma^2/\sigma_U^2, \lambda_V = \sigma^2/\sigma_V^2$ and $\lambda_\mu = \sigma^2/\sigma_\mu^2$ and a multiplicative constant $1/\sigma^2$ is dropped in all three equations.

To learn model parameters, we alternatively update $U, V$ and $\mu$ using the gradient algorithm with a learning rate $\eta$ by maximizing the log-likelihood. First we update $U, V$ by

$$U_i \leftarrow U_i + \eta \frac{\partial \mathcal{L}}{\partial U_i}, \qquad V_j \leftarrow V_j + \eta \frac{\partial \mathcal{L}}{\partial V_j}. \quad (19)$$

Then using the updated $U, V$, we update $\mu_l$ by

$$\mu_l \leftarrow \mu_l + \eta \frac{\partial \mathcal{L}}{\partial \mu_l}. \quad (20)$$

Similar to PMF [22], we linearly map the rating values in $[1, D]$ to $[0, 1]$ and pass $U_i^T V_j$ through the sigmoid function as defined in Eq. (13). To avoid cluttered notations, we drop all the logistic function in our derivation process. After obtained the trained model, we convert the expected value, $g(U_i^T V_j)$, back to the scale of 1 to $D$ and set it as the predicted score of user $i$'s rating on item $j$.

### 3.5 Context aware response model

In real-world recommender systems, the probability of an item being rated may not only depend on users' rating score. Many factors affect the response probabilities. For example, in a movie rating system, some popular movies such as Titanic, Avatar, may have much higher probability of being rated than a mediocre movie. Moreover, the features of users and items may contain group structure [30]. One may argue this might be caused by the higher inspection rate, i.e., it is likely that a reputable movie is being watched more than an obscure one. Nevertheless, it still makes sense that some items may have higher chance of receiving a rating due to the high quality that a user will not hesitate to rate it. In addition, different user may have distinct rating habits. Some users might be more willing to provide ratings in order to get high quality recommendation. This is supported by the fact that the number of ratings received from different users can differ wildly in real-world deployed recommender systems.

To capture such factors, we generalize the rating dominant response model by including both item features and user features. To keep the model tractable and efficient, we introduce a linear combination of the item features, user features and a constant related to the rating scores and pass it through the logistic function to model the response probability,

$$\mu_{ijk} = \frac{1}{1+\exp(-(\delta_k + U_i^T \boldsymbol{\theta}_U + V_j^T \boldsymbol{\theta}_V))}. \quad (21)$$

We refer to Eq. (21) as context-aware response model, in which the response probability is on a per-user-item-rating basis. More sophisticated relationship definition can be referred to [28, 29, 31]. The PMF integrated with the context-aware response model is named RAPMF-c. Note that by setting $\boldsymbol{\theta}_U$ and $\boldsymbol{\theta}_V$ to zero, we can recover the rating dominant response model in Eq. (13).

The log-likelihood of RAPMF-c is in the same structure as RAPMF-r. We only need to substitute $\mu_k$ in Eq. (15) by $\mu_{ijk}$ defined in Eq. (21). Similarly, the gradients of $\mathcal{L}$ with respect to $U_i$ and $V_j$ are

$$\frac{\partial \mathcal{L}}{\partial U_i} = \beta \sum_{j=1}^{M} \frac{\sum_{k=1}^{D} t_{Ukij}\mathcal{N}(k|U^T V, \sigma^2)}{\sum_{k=1}^{D} \alpha_{kij}\mathcal{N}(k|U^T V, \sigma^2)}$$
$$- \sum_{j=1}^{M}(U_i^T V_j - x_{ij})[r_{ij}=1]V_j - \lambda_U U_i, \quad (22)$$

$$\frac{\partial \mathcal{L}}{\partial V_j} = -\beta \sum_{i=1}^{N} \frac{\sum_{k=1}^{D} t_{Vkij}\mathcal{N}(k|U^T V, \sigma^2)}{\sum_{k=1}^{D} \alpha_{kij}\mathcal{N}(k|U^T V, \sigma^2)}$$
$$- \sum_{j=1}^{M}(U_i^T V_j - x_{ij})[r_{ij}=1]U_i - \lambda_V V_j, \quad (23)$$

where the $t_{Ukij}$ and $t_{Vkij}$ is defined as following

$$t_{Ukij} = g'(\mu_{kij})(-1)^{[r_{ij}=0]}\boldsymbol{\theta}_U - \alpha_{kij}(U_i^T V_j - k)V_j, \quad (24)$$

$$t_{Vkij} = g'(\mu_{kij})(-1)^{[r_{ij}=0]}\boldsymbol{\theta}_V - \alpha_{kij}(U_i^T V_j - k)U_i. \quad (25)$$

Correspondingly, the gradients of $\mathcal{L}$ with respect to $\delta_l$, $\boldsymbol{\theta}_U$ and $\boldsymbol{\theta}_V$ are

$$\frac{\partial \mathcal{L}}{\partial \delta_l} = \sum_{i=1}^{N}\sum_{j=1}^{M} \frac{\mathcal{N}(l|U^T V, \sigma^2)g'(\mu_{kij})(-1)^{[r_{ij}=0]}}{\sum_{k=1}^{D} \alpha_{kij}\mathcal{N}(k|U^T V, \sigma^2)}$$
$$-\lambda_\mu \delta_l, \quad (26)$$

$$\frac{\partial \mathcal{L}}{\partial \boldsymbol{\theta}_U} = \sum_{i=1}^{N}\sum_{j=1}^{M} \frac{\sum_{k=1}^{D}\mathcal{N}(k|U^T V, \sigma^2)g'(\mu_{kij})(-1)^{[r_{ij}=0]}U_i}{\sum_{k=1}^{D} \alpha_{kij}\mathcal{N}(k|U^T V, \sigma^2)}$$
$$-\lambda_\mu \boldsymbol{\theta}_U, \quad (27)$$

$$\frac{\partial \mathcal{L}}{\partial \boldsymbol{\theta}_V} = \sum_{i=1}^{N}\sum_{j=1}^{M} \frac{\sum_{k=1}^{D}\mathcal{N}(k|U^T V, \sigma^2)g'(\mu_{kij})(-1)^{[r_{ij}=0]}V_j}{\sum_{k=1}^{D} \alpha_{kij}\mathcal{N}(k|U^T V, \sigma^2)}$$
$$-\lambda_\mu \boldsymbol{\theta}_V. \quad (28)$$

To learn RAPMF-c, we adopt the alternatively updating scheme to maximize the log-likelihood, where the updating rules of $U$ and $V$ are the same as those in Eq. (19). After updating $U$ and $V$, we update $\delta_l$, $\boldsymbol{\theta}_U$ and $\boldsymbol{\theta}_V$ by

$$\boldsymbol{\vartheta} \leftarrow \boldsymbol{\vartheta} + \eta \frac{\partial \mathcal{L}}{\partial \boldsymbol{\vartheta}},$$

where $\boldsymbol{\vartheta}$ is replaced by $\delta_l$, $\boldsymbol{\theta}_U$ and $\boldsymbol{\theta}_V$, respectively.

### 3.6 Complexity and Parallelization

The training complexity of RAPMF, $O(MN)$, can be quite time consuming compared with the PMF, which is linear in number of observations, $O(|\mathcal{Q}|)$. However, we argue that the time spent on training is worthy since it can boost the model performance. More importantly, the prediction complexity of RAPMF is the same as PMF, $O(K)$, which can be taken as a constant time given a moderate sized $K$. Since the training procedure can be performed offline, RAPMF can accommodate the hard response time constraint in real-world deployed recommender systems due to the succinct prediction cost.

In addition, RAPMF can be speedup by parallelization. The intensive computation cost, calculating the gradients, can be decoupled and distributed to a cluster of computers. It is also possible to use online learning to speed up the training process [12].

## 4 Experiments and Results

We conduct empirical evaluation to compare the performance of PMF [22], CPT-v [16], Logit-vd [15], and our RAPMF. We try to answer the following questions:

1. How to collect data with benchmark response patterns to evaluate the models fairly?

2. How to design experiment protocols to evaluate the performance the models with and without response models fairly?
3. How the compared models perform on the collected data?
4. How the parameters, $\beta$ and $\lambda$, affect the performance of RAPMF?

Section 4.1-4.4 answer the above questions, respectively.

## 4.1 Datasets

We conduct our empirical analysis on two datasets: a synthetic dataset and a real-world dataset, the *Yahoo! Music ratings for User Selected and Randomly Selected songs, version 1.0* (Yahoo dataset)[1].

**Synthetic dataset.** The data generation process consists of two steps: generating full rating matrix and generating response matrix. To generate the full rating matrix, we first generate the latent user features and item features from zero-mean spherical Gaussian as follows:

$$U_i \sim \mathcal{N}(\mathbf{0}_K, \sigma_U^2 \mathbf{I}_K), \quad V_j \sim \mathcal{N}(\mathbf{0}_K, \sigma_V^2 \mathbf{I}_K),$$

where $i = 1, \ldots, N$, $j = 1, \ldots, M$, $\mathbf{0}_K$ is a $K$-dimensional vector with each element being 0 and $\mathbf{I}_K$ is the $K \times K$ identity matrix. The full rating matrix $X$ is then obtained by re-scale the sigmoid value of $U^T V$ to 1 to $D$ by $X_{ij} = \lceil g(U_i^T V_j) \times D \rceil$.

To generate the response matrix $R$, we first set the inspection probability of a user inspecting an item, $P_{inspect}$. Then, the partitioning of inspected ratings and un-inspected ratings are done by the Bernoulli trails with success probability $P_{inspect}$. For all the inspected ratings, we model their response probability by a Bernoulli distribution with the success probability $P_k$, where $k \in \{1, 2, \ldots, D\}$. Table 2 summarizes the parameters used for generating the synthetic dataset. The parameters are selected so that they can faithfully simulate real users' ratings and response behaviors. The rating probabilities $P_k$ are chosen according to Fig. 1. To minimize the effect of randomness, we generate the dataset independently 10 times and report the average result in the following. On average, we provide about 3.3% of the full matrix as training set, around 3.4% as testing set for traditional protocol, around 17.3% as testing set for adversarial protocol and all the remaining 80% as testing set for realistic protocol.

**Yahoo dataset.** It provides a unique opportunity to investigate the response ignorance problem. The dataset contains 311,704 *training ratings* collected from 15,400 users on 1,000 songs during the normal interaction between the users and the Yahoo! Music system, with at least 10 ratings for each user. During a survey conducted by Yahoo! Research, exactly 10 songs randomly selected from these 1,000 songs are presented to the user to listen and rate. In total there are 5,400 users participated this survey and these 54,000 ratings are the *testing ratings*.

[1]http://webscope.sandbox.yahoo.com

Table 2: Parameters for generating the synthetic dataset.

| $N$ | $M$ | $D$ | $K$ | $P_{inspect}$ |
|---|---|---|---|---|
| 1000 | 1000 | 5 | 5 | 0.2 |
| $P_1$ | $P_2$ | $P_3$ | $P_4$ | $P_5$ |
| 0.073 | 0.068 | 0.163 | 0.308 | 0.931 |

## 4.2 Setup and Evaluation Metrics

In a real-world deployed recommender system, the status of an item given a user follows exactly one of the three types: *un-inspected*, *inspected-unrated*, and *inspected-rated*. Traditional collaborative filtering approaches separate the inspected-rated data into training set and test set and evaluate the model on the test set. Since both the training set and the test set belongs to the inspected-rated type, their rating distributions are the same. Thus, the traditional evaluation scheme may hide the significance of the ignoring response model. In the experiment, we first investigate two existing experimental protocols:

- **Traditional protocol:** Both the training set and the test set are randomly selected from inspected-rated items together with their rating users and assigned scores. This is exactly the traditional experiment protocol [22].
- **Realistic protocol:** The training set is randomly selected from inspected-rated items, but the test set is randomly selected from un-inspected items. This is an experimental protocol adopted in [15, 16]. This protocol captures the ultimate goal of a recommender system, i.e., recommending un-inspected items to potential users who are interested.

Moreover, we will investigate a new experimental protocol:

- **Adversarial protocol:** The training set is randomly selected from inspected-rated items, but the test set is randomly selected from inspected-unrated items. This setting tests the model's performance when the distribution of the training set and test are very divergent. It can reveal the property of the model in some real-world cases where most of inspected-rated items receive very high scores, while those inspected-unrated items have low scores. This setting also demonstrates the model performance when we have an adversary that manipulates the responses.

For the synthetic dataset, we use the same training set for all three protocols. For various protocols, different test set can reveal different properties of the PMF with and without the response models. We report the average performance on the 10 independently generated datasets.

For Yahoo dataset, we use only traditional and realistic protocol and do not evaluate the adversarial protocol due to the missing of necessary inspection information. For the traditional protocol, we perform 10 fold cross validation on the training ratings. For the realistic protocol, we train the model using training ratings and test on the testing ratings. We perform the experiment 10 times and report the average results.

In the experiment, we use Root Mean Square Error (RMSE) to evaluate the performance of various approaches [16, 22, 27], i.e., RMSE = $\sqrt{1/|\mathcal{T}|\sum_{(i,j,x)\in\mathcal{T}}(\hat{x}_{ij}-x)^2}$, where $\mathcal{T}$ is the set of $(i,j,x)$ triplets reserved for testing and $\hat{x}_{ij}$ is model prediction for user $i$'s rating on item $j$.

### 4.3 Model Comparison

For both the synthetic dataset and Yahoo dataset, we randomly select 10% of the testing ratings from realistic protocol as validation set to tune the parameters (more advanced techniques can be referred to [20]):

- $\lambda_U$ and $\lambda_V$: They are tuned by the grid search scheme, i.e., first selecting from $\{10^{-3}, 10^{-2}, 10^{-1}, 10^0, 10, 10^2\}$, respectively. We then fine-tune the range to achieve the best performance of PMF; see an example in Fig. 3(a).
- $\beta$: We first fix the optimal $\lambda_U$ and $\lambda_V$ obtained from PMF, we then tune it in $\{0.0, 10^{-3}, 10^{-2}, 10^{-1}, 1.0\}$ and fine-tune it further for RAPMF-r; see an example in Fig. 3(b).
- $\lambda_\mu$: As shown later in Fig. 3(c), this parameter is insensitive on a large range.

These parameters are then used across different protocols. The hyper-parameters used for CPT-v and Logit-vd follow the settings used in [15]. We choose $K = 5$ as the latent dimension size for all the experiments. All the models are trained using 500 iterations. According to our experience, the change in performance after 200 iterations is negligible.

Figure 2 shows the results of various models' performance under different protocols on both the synthetic and Yahoo datasets. Figure 2(a) shows the results on the synthetic datasets. We see that PMF performs best under traditional protocol. This is expected because under the traditional setting, the testing set and the training set have exactly the same distribution. The response model does not help. However, under realistic and adversarial protocol, the proposed RAPMF-r outperforms PMF by 5.5% and 9.2%, with 95% confidence level on the paired $t$-test, respectively. The RAPMF-c performs slightly worse than RAPMF-r. This is probably due to the reason that we does not take the user and item features into account when generating the dataset.

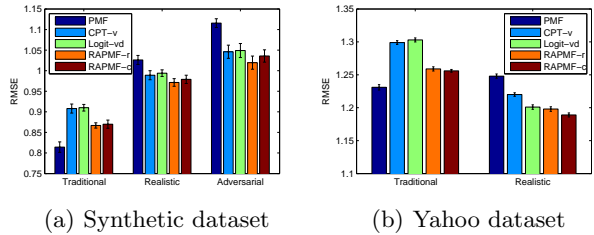

(a) Synthetic dataset    (b) Yahoo dataset

Figure 2: Relative performance of various models. Smaller value indicates a better model performance.

More importantly, the learned rating probability for a typical run of RAPMF-r is [0.0125, 0.0124, 0.0155, 0.0267, 0.105]. Comparing this with the parameter used in Table 2 when generating the data, we see that although RAPMF cannot recover the rating probabilities exactly, the overall trend is captured quite precisely. This explains the significant performance boost in realistic and adversarial protocol.

Figure 2(b) shows the results on the Yahoo dataset. Again, PMF attains the best performance under traditional protocol. Under realistic protocol, the RAPMF-r and RAPMF-c outperforms PMF by 4.1% and 4.9%, with 95% confidence level on the paired $t$-test, respectively. The performance gain is slightly less than that in the synthetic dataset, probably due to the reason that the rating probability in real-world dataset is not as dynamic as the value we choose in Table 2. The performance boost gained from context-awareness is not as significant as we have expected. This result hints that the rating value might impact a user's decision on whether to rate the item more than the user and item features.

### 4.4 Sensitivity Analysis

In the following, we investigate how the model parameters affect the performance of RAPMF. All the sensitivity analysis is done under the realistic setting in one-trial of the generated synthetic dataset.

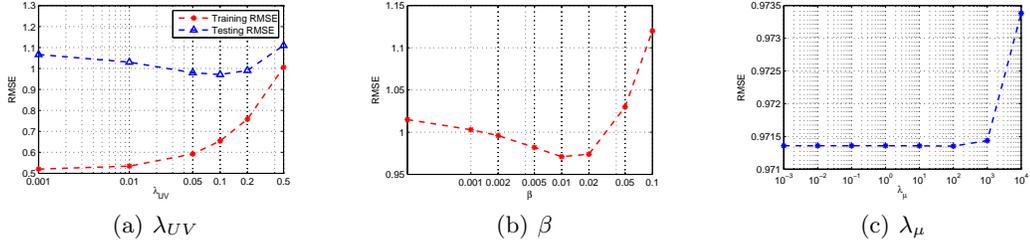

Figure 3: Sensitivity analysis of hyper-parameters on RAPMF on one-trial test.

#### 4.4.1 Impact of $\beta$

The faith parameter $\beta$ is arguably the most important parameter in our RAPMF model. As has been discussed in Section 3, we adopt a soft-margin approximation of the ideal hard-margin response model. When making this approximation, we unavoidably introduce some bias in the response model because we cannot fully recover the true $U$ and $V$. Hence, we introduce $\beta$ to control the weight of the response model.

Figure 3(b) plots the performance of RAPMF versus $\beta$ on the logarithmic scale. When $\beta = 0$, RAPMF fall back to PMF, whose performance is 1.015, corresponding to the most left point in Fig. 3(b). Clearly, by incorporating an explicit response model, RAPMF is able to beat PMF by a large margin (nearly 5%) by using a proper $\beta$ value. However, if we use too large a $\beta$, we quickly lose the boost provided by the response model. An observation of the experiment is that when $\beta$ is too large, the model does not converge. This is because the model training starts from randomly initialized $U$ and $V$, a large weight on the response model will pull the model away from the true model and cause divergence.

#### 4.4.2 Impact of $\lambda$'s

The regularization parameters $\lambda$ are placed on $U$, $V$ and $\mu$. Since in the dataset, users and items are symmetric, we use the same regularization parameter $\lambda_{UV}$ for $U$ and $V$ and use another parameter $\lambda_\mu$ to control $\mu$.

Figure 3(a) shows the impact of $\lambda_{UV}$ on the performance of RAPMF. When $\lambda_{UV}$ is very small, although the RAPMF is able to fit the training data very well, it does not generalize well to the test set. This is a sign of over-fitting. As $\lambda_{UV}$ becomes larger, which limits the norms of $U$ and $V$, the training RMSE increases but the test RMSE decreases gradually. However, after a turning point, both the training RMSE and test RMSE start to increase. This is when the regularization is too stringent that it hinders the proper fitting of the model.

Figure 3(c) shows the impact of $\lambda_\mu$ on the model performance. As we can see, it is basically a straight line in a large range from $10^{-3}$ to $10^4$, while the RMSE is changed only from 0.9714 to 0.9734, a very small scale. The effect of $\lambda_\mu$ is inappreciable. This is probably due to the fact that $\mu$ is a parameter in $D$-dimension ($D$=5) and no significant over-fitting can occur.

## 5 Conclusion and Future Work

In this paper, we propose two response models, rating dominant and context-aware response models, to capture users' response patterns. Further, we unify the response models with one of famous collaborative filtering model-based methods, the Probabilistic Matrix Factorization, to establish the Response Aware Probabilistic Matrix Factorization framework (RAPMF). The RAPMF also generalizes PMF as its special case. Empirically, we verify the performance of RAPMF under carefully designed experimental protocols and show that RAPMF performs best when it tries to fulfill the ultimate goal of real-world recommender systems, i.e., recommending items to those who may be interested in. The empirical evaluation demonstrates the potential of our RAPMF model in real-world recommender system deployment.

There are several interesting directions worthy of considering for future study. One direction is to study how to model the response when the response patterns are hidden. The second fascinating avenue is to study how to speed up our RAPMF through parallelization, online learning, or sampling techniques. The third direction is to design a smart way to efficiently tune the hyper-parameters or to design the learning scheme to automatically learn the model parameters.

## 6 Acknowledgments

The work described in this paper was fully supported by the Shenzhen Major Basic Research Program (Project No. JC201104220300A) and the Research Grants Council of the Hong Kong Special Administrative Region, China (Project Nos. CUHK413210, CUHK415311 and N_CUHK405/11).